%% file: paper.tex
\documentclass[]{bytedance_seed}

\usepackage[toc,page,header]{appendix}
\usepackage{amsfonts}

\usepackage{minitoc}

\title{SeedEdit 3.0: Fast and High-Quality Generative Image Editing}

\author[*\dagger]{Peng Wang}
\author[*]{Yichun Shi}
\author[]{Xiaochen Lian} 
\author[]{Zhonghua Zhai}
\author[]{Xin Xia}
\author[]{Xuefeng Xiao}
\author[]{Weilin Huang}
\author[]{Jianchao Yang}

\affiliation[]{ByteDance Seed}
\contribution[*]{Equal Contribution}
\contribution[\dagger]{Project Lead}

\abstract{
We introduce SeedEdit 3.0, in companion with our T2I model Seedream 3.0~\cite{seedream3.0}, which significantly improves over our previous version~\citep{shi2024seededit} in both aspects of edit instruction following and image content (e.g., ID/IP) preservation on real image inputs. Additional to model upgrading with T2I, in this report, we present several key improvements. 
First, we develop an enhanced data curation pipeline with a meta-info paradigm and meta-info embedding strategy that help mix images from multiple data sources. This allows us to scale editing data effectively, and meta information is helpfult to connect VLM with diffusion model more closely. 
Second, we introduce a joint learning pipeline for computing a diffusion loss and a reward loss. 
Finally, we evaluate SeedEdit3.0 on our testing benchmarks, for real image editing, where it achieves a best trade-off between multiple aspects, yielding a high usability rate of 56.1\%, compared to SeedEdit 1.6~\cite{shi2024seededit} (38.4\%), GPT4o~\cite{hurst2024gpt} (37.1\%) and Gemini 2.0~\cite{gemini2.0} (30.3\%). SeedEdit 3.0 will be online in Jimeng~\footnote{https://jimeng.jianying.com/}, Doubao\footnote{https://www.doubao.com/} and other Bytedance Apps. }

\checkdata[Page]{\url{https://seed.bytedance.com/tech/seededit}}

\begin{document}
\maketitle
\input{sections/introduction}
\input{sections/relatedwork}

\input{sections/approach}

\input{sections/experiments}

\bibliographystyle{plainnat}
\bibliography{main}

\clearpage

\beginappendix

\input{sections/appendix}

\end{document}

%% file: sections/introduction.tex
\section{Introduction}

\begin{figure}[t!]
    \includegraphics[width=1.0\linewidth]{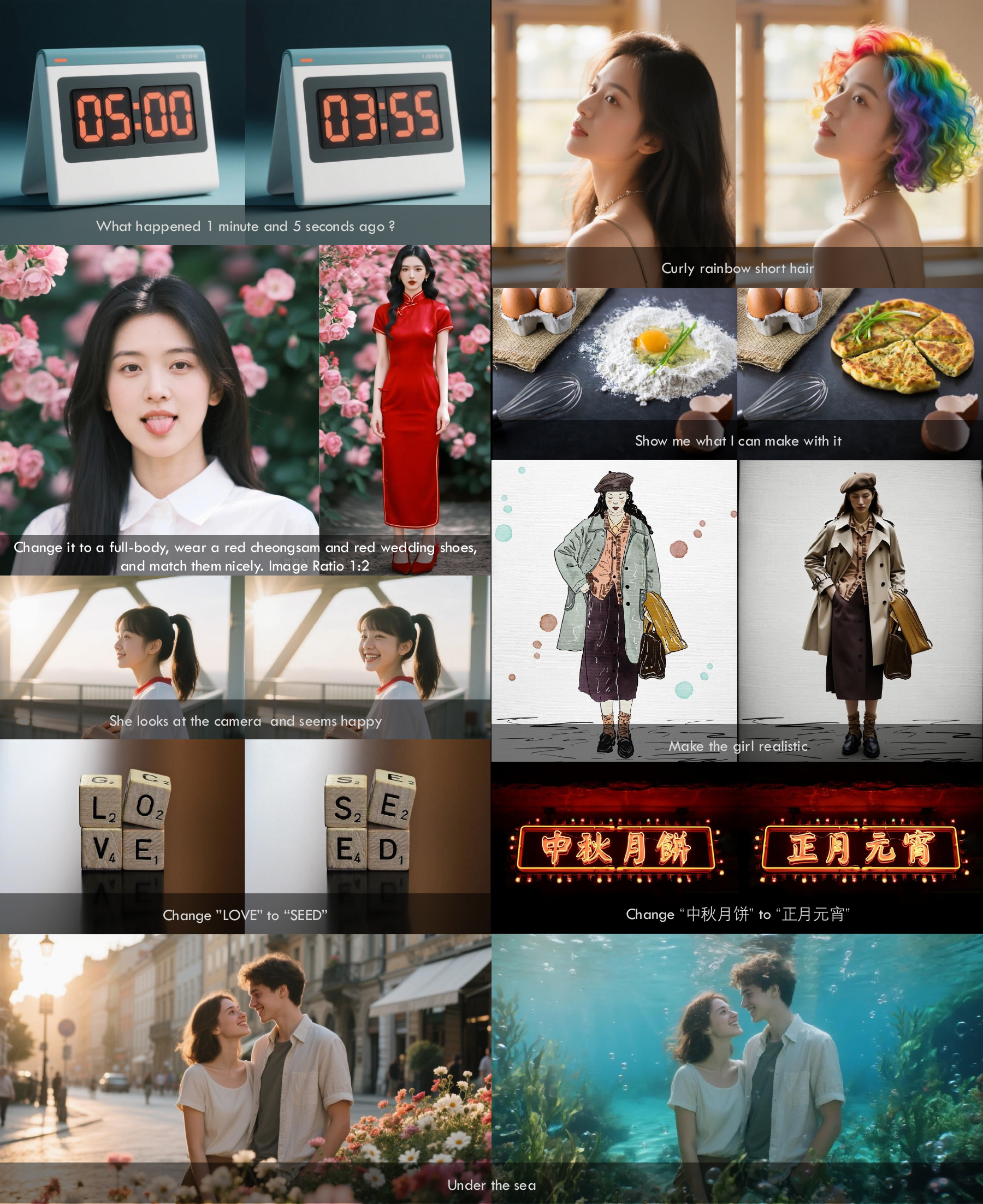}
    \caption{Example images edited by SeedEdit3.0 with real and generated images as input, which provides high detail in ID preservation and strong edit intention understanding.}
    \label{fig:teaser}
\end{figure}

As the size of the Text-to-Image (T2I) generation model increases, the performance of the model has become increasingly dependent on the quantity and quality of the available data. This is particularly important for tasks involving instructive editing of real images, which is the focus of this work. 
% With the increasing size of models for Text-to-Image (T2I) generation, the performance of the final model has become heavily dependent on the quality and quantity of data available. This is particularly true for tasks involving instructive editing given real images as inputs, which is the focus of this work.   

Previous works~\cite{hui2024hqedit,zhao2024ultraedit,brooks2023instructpix2pix} have introduced various methods for data generation, encompassing both real images and synthetic data, resulting in numerous datasets of varying quality. Some datasets feature high-quality images with minimal expert editing~\cite{yu2024promptfixpromptfixphoto}, while others~\cite{brooks2023instructpix2pix} exhibit substantial and diverse changes accompanied by noise. Therefore, determining the optimal approach to leverage different datasets and extract the best components from each in a unified model is crucial to develop a robust, general-purpose instructive editing model.
% Previous works~\cite{hui2024hqedit,zhao2024ultraedit,brooks2023instructpix2pix} have introduced various methods for data generation, encompassing both real images and synthetic data, resulting in numerous datasets of varying quality. Certain datasets feature high-quality images that contain minimal expert editing~\cite{yu2024promptfixpromptfixphoto}. In contrast, other datasets~\cite{brooks2023instructpix2pix} exhibit substantial diverse changes which accompanied by noise. Thus, determining the optimal approach to leverage different datasets and extract the best components from each into a unified model is crucial to develop a robust general instructive editing model. 

One potential strategy is to utilize a quality score derived from a Vision-Language Model (VLM) for each data point and subsequently incorporate these scores as conditions within the diffusion framework. During inference, the highest quality score is used as the primary condition, as proposed in methods like HIVE~\cite{zhang2024hive}. However, due to the feature gap between VLMs and diffusion models, this quality score can introduce particular biases from the VLM, leading to suboptimal results in the diffusion model. To address this, in this paper we propose a meta-information strategy that annotates data with labels or captions of multiple granularity, which significantly helps the diffusion model distinguish different datasets and find the best trade-off of mixing the datasets.
% One potential strategy is to utilize a quality score derived from a Vision Language Model (VLM) for each data point, subsequently merging these scores as conditions in the diffusion framework. During the inference phase, we employ the highest quality score as the primary condition. As proposed in methods like HIVE~\cite{zhang2024hive}.  However, due to the existing feature gap between VLMs and diffusion models, this quality score can introduce particular biases from the VLM, leading to suboptimal results in the diffusion model. Therefore, in this work, we introduce a meta-info strategy that labels data with labels/captions of multi-granularity, which significantly helps the diffusion model distinguish different datasets and find the best trade-off of mixing the datasets. 

For data curation, we proposed multiple data sources, including using our internal T2I~\cite{seedream3.0} and SeedVLM~\cite{seed15vl} which will be introduced in Sec.~\ref{subsec:data}. Then, we are able to generate images with resolutions greater than 1024$\times$1024 with rich captions, facilitating high-resolution image editing and understanding while preserving intricate details from the input images, such as facial identification and hair textures. 
Last but not least, to enhance the quality of certain preferable features--such as face alignment, text rendering--we additionally develop several specialized models that can be jointly trained with diffusion models. In Fig.~\ref{fig:teaser}, we present several examples demonstrating our model's ability to handle real images, by following complex instructions and preserving details. 

\begin{figure}[t]
\centering
\begin{subfigure}[b]{0.59\textwidth}
\includegraphics[width=\linewidth]{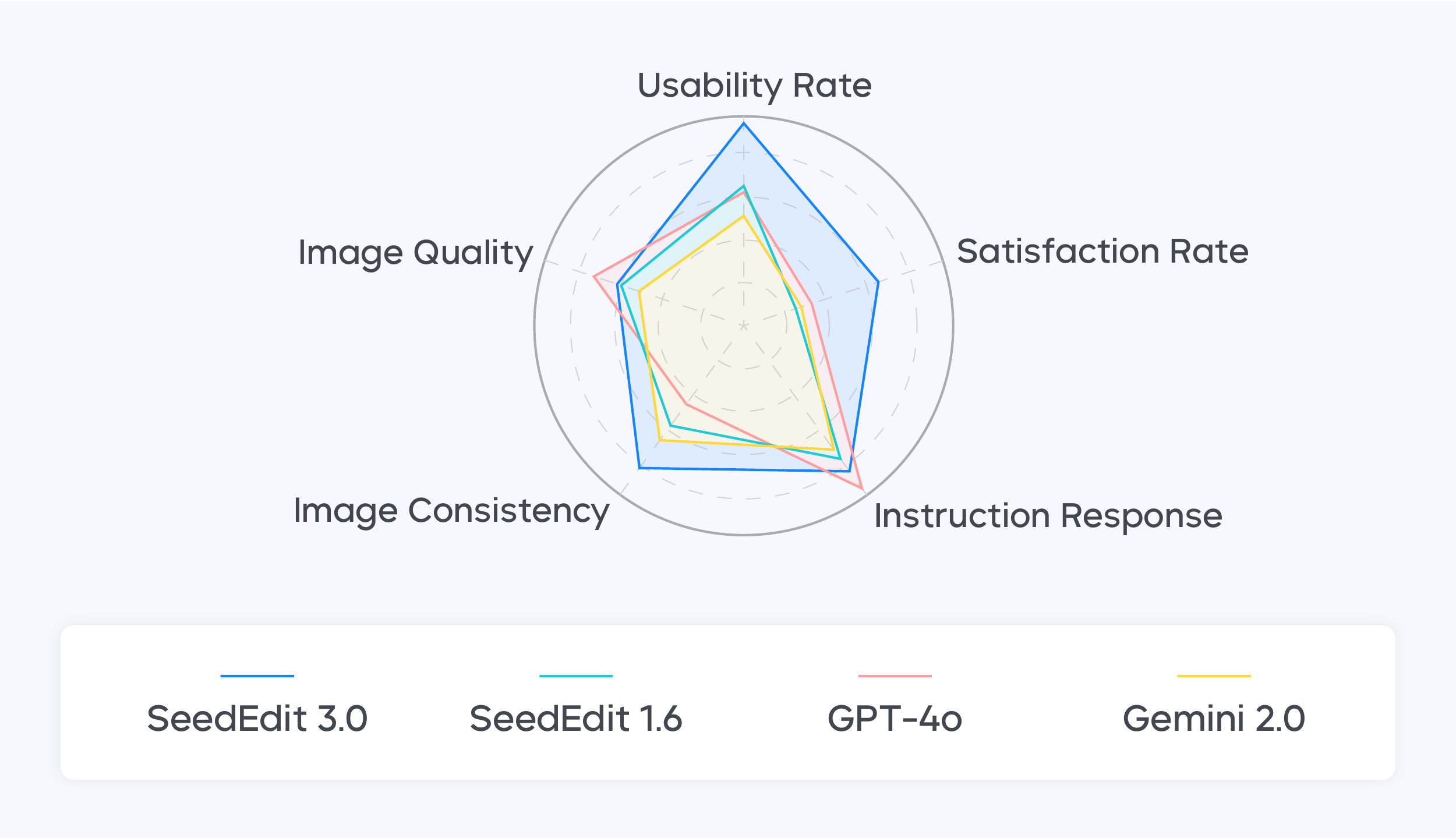}\hfill
\caption{Performance}
\end{subfigure}\hfill
\begin{subfigure}[b]{0.4\textwidth}
\includegraphics[width=\linewidth]{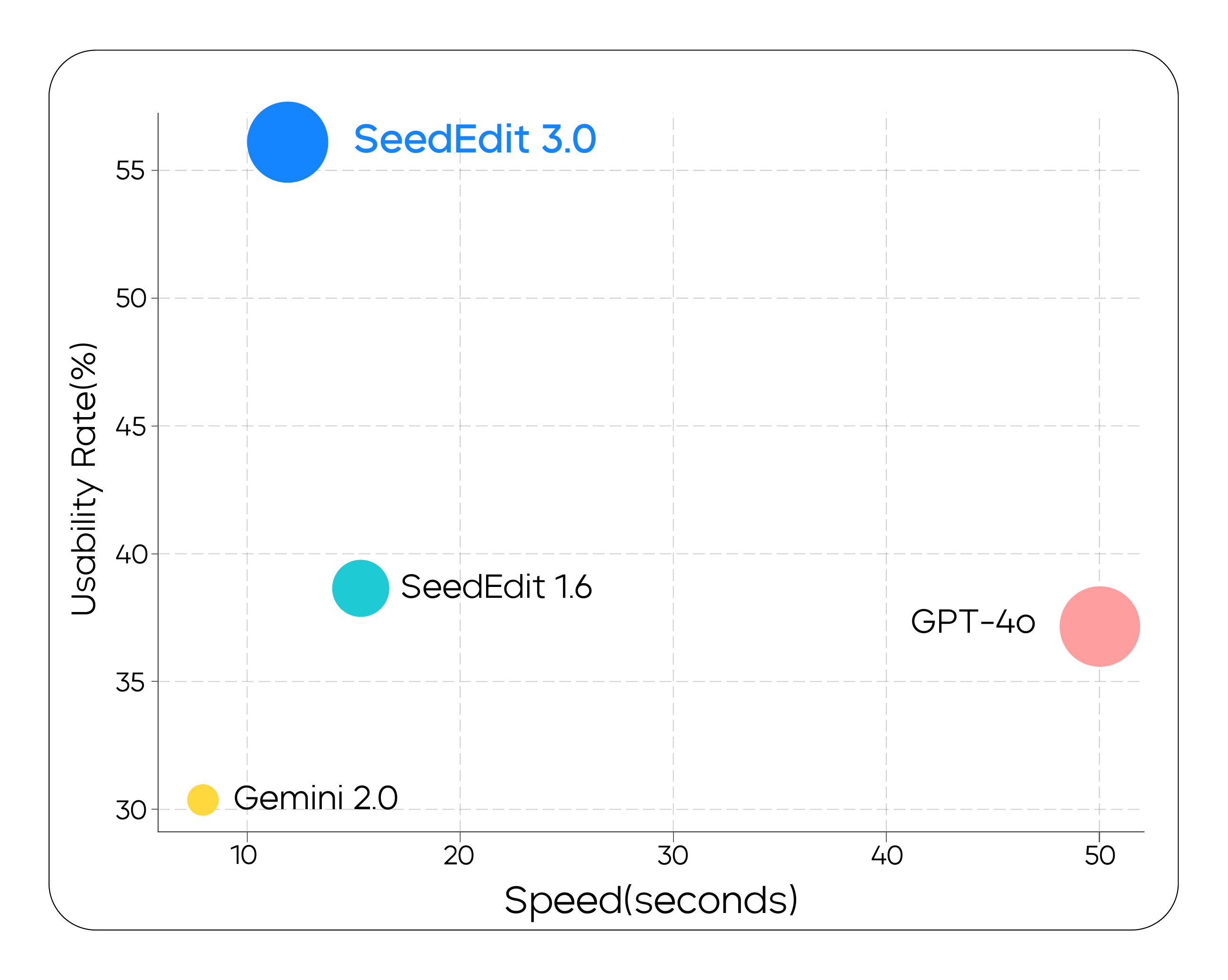}
\caption{Speed}
\end{subfigure}
\caption{Overview of SeedEdit3.0 human evaluation.  \textbf{Left} Spider Graph of ours vs. other methods on various metrics. Details in Sec.~\ref{sec:experiment}.
\textbf{Right}: Speed and usability rate comparison. Dot size represent roughly the model size. We illustrate hypothesized size of GPT4o and Gemini2.0 based on their speed. For SeedEdit, although the model size increases, the pipeline is simplified, so the speed improves as well.}
\label{fig:performance}
\end{figure}

We compare our results with those of the state-of-the-art (SoTA) products, such as our online version SeedEdit1.6~\cite{shi2024seededit}, Gemini 2.0~\cite{gemini2.0} and GPT-4o~\cite{hurst2024gpt}, and show that our method achieves the best trade-off in terms of human preference, demonstrating its effectiveness. 
In Fig.~\ref{fig:performance}, we show the comparison of human evaluation against other SoTA commercial models using an internal evaluation benchmark that includes real images and more diverse instructions than existing public benchmarks. Although GPT-4o has the best instruction response, SeedEdit3.0 achieves the best trade-off among multiple evaluation metrics, including editing instruction following, content preservation and image quality. In addition, our model is significantly faster than GPT-4o (e.g., 15 s vs. 50 s per query).

% In summary, SeedEdit 3.0 ahcieves the best trade-off of editing instruction following, marked as "usability rate", while GPT-4o has the best instruction response. In addition, our model is also substantially faster than GPT-4o (10 s vs. 50 s per query).

% Show the results and evaluations; in Fig.1  Illustrate data fusion pipeline in Fig. 2; 

%% file: sections/relatedwork.tex
\section{Related Work}
% \peng{Xiaochen, please help this part}
We briefly review a few recent methods for instructive editing, covering the two most important topics: diffusion model methods and data set creation.

\paragraph{\textbf{Instruct Editing Methods}}
% \peng{xiaochen data : training free methods can be simply referred, major mention some training methods such as adobe's, step1x, gpt and gemini etc. }

Existing methods using diffusion models for image editing can be primarily categorized into two groups: training-free approaches and training approaches. The training-free method controls the generation of images in the denoising process, by inverting the diffusion process~\cite{mokady2023null,nie2023blessing,feng2025dit4edit,lin2024schedule} and attention crontrol~\cite{hertz2022prompt,cao2023masactrl,tumanyan2023plug}. Although fast and low-cost, they all suffer from inferior content preservation and low editing accuracy, e.g., inconsistency with either the input image or the target descriptions. 

To achieve the best editing quality, it is widely acknowledged that retraining a diffusion model is necessary. Early training-based approaches~\cite{wang2023instructedit, brooks2023instructpix2pix, zhang2023magicbrush, shi2024seededit} train diffusion-based editing models on synthesized image editing datasets. Later works focus on novel model architectures for better instruction-image interaction~\cite{huang2024smartedit, yang2024mastering, yu2024anyedit, zhao2024ultraedit}. Recently, unified image generation and editing framework has attracted more and more attention. OmniGen~\cite{xiao2024omnigen}, Transfusion~\cite{zhou2024transfusion},  and Mogao~\cite{liao2025mogao} jointly model text and images within a single transformer to achieve unified representations. DreamEngine~\cite{chen2025multimodal}, MetaQueries~\cite{pan2025transfer} and Step1X-Edit~\cite{liu2025step1x} connect the text and image latent features of Multimodal LLM (MLLM) to the diffusion decoder, leveraging MLLM's strong capabilities in understanding and reasoning. By joint vision-language training, Gemini 2.0~\cite{gemini2.0} and GPT-4o~\cite{hurst2024gpt} have demonstrated strong performance in following instruction and generating consistent images. In this paper, we will compare model with SeedEdit, GPT-4o and Gemini on real image editing tasks.

\paragraph{\textbf{Dataset Creation}} 
% \peng{xiaochen data : omniedit, anyedit, instrutive move etc. }

One of the major challenges in training an instruction-based image editing model is the lack of large-scale datasets that have high-quality image editing pairs with corresponding instructions. Early approaches like Magicbrush~\cite{zhang2023magicbrush} construct datasets by manually labeling image pairs, which is not scalable and can hardly cover all types of image editing. InstructPix2Pix~\cite{brooks2023instructpix2pix} and HIVE~\cite{zhang2024hive} leverage GPT-3~\cite{brown2020language} and Prompt2Prompt~\cite{hertz2022prompt} to generate image editing pairs. HQ-EDIT~\cite{hui2024hqedit} and UltraEdit~\cite{zhao2024ultraedit} further push the quality of this data synthesis pipeline by using more powerful foundation models such as GPT-4V~\cite{hurst2024gpt} and DALL-E 3~\cite{dalle3}. To ensure the diversity and quality of the synthesized data, Seed-Edit~\cite{shi2024seededit} combines different regeneration techniques and sampling hyperparameters and applies importance sampling to obtain diverse and high-quality training examples. Synthesized data has a strong bias towards the underlying generative models. To handle real images, \cite{wei2024omniedit,sheynin2024emu} train multiple expert models, each specializing in a different editing task, to generate a large, high-resolution, multi-aspect ratio dataset. Our data curation pipeline, as described in Sect.~\ref{subsec:data}, inherits all the merits of existing pipelines, e.g., including both synthesized and real images of high quality, multiple sources, and different sizes, utilizing LLM/VLM to enrich the editing instructions and to ensure precise alignment between the paired images and their corresponding instructions, designing a reliable data assessment process to control the data quality, etc.

% Existing methods using diffusion models for image editing can be primarily categorized into three groups: training-free approaches, inference-time fine-tuning approaches, and data-driven training approaches. 

% Firstly, the training-free method controls image generation in the denoising process. Text-based methods refine input texts [] or enhance text embeddings [] to enable conceptual modifications and user instruction following. \cite{mokady2023null}[null-text, SDE-Drag, DiT4Edit, Schedule Your Edit] apply solver inversion (e.g., DDIM and DPM-solver inversion) to construct the noise latent space of input images and regenerate the target images conditioned on the text prompts. Attention control methods [Prompt-to-Prompt, masactrl, Plug-and-play] manipulate the attention features during the reverse diffusion steps to achieve the editing goals and content consistency. Although fast and low-cost, training-free methods suffer from inferior content preservation and low editing accuracy, e.g., inconsistency with either the input image or the target descriptions.

% Step1X-Edit
% Gemini 2.0
% Janus: Decoupling visual encoding for unified multimodal understanding and generation.
% Janus-pro: Unified multimodal understanding and generation with data and model scaling
%  Show-o: One single transformer to unify multimodal understanding and generation

%% file: sections/approach.tex
\section{Approach}

\subsection{Data Curation}
\label{subsec:data}
\begin{figure}[t]
    \centering
    \includegraphics[width=\linewidth]{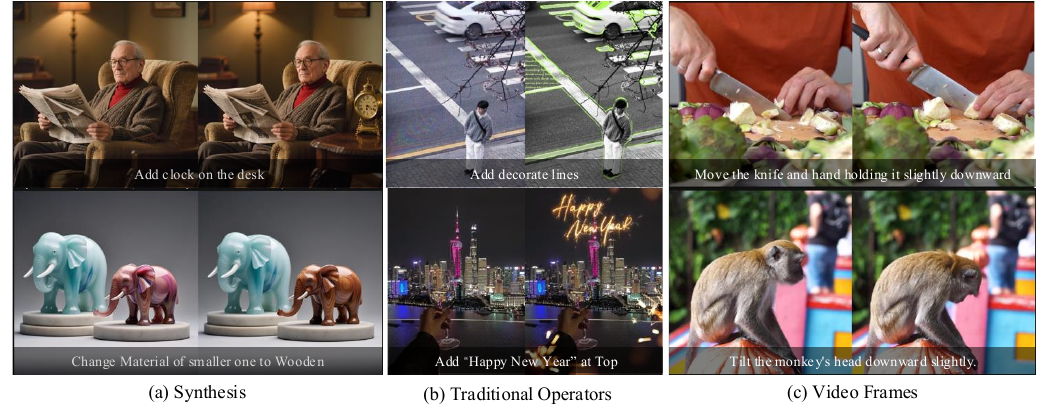}
    \caption{Few data examples from our data curation pipeline. Each example, we will have task label, optimized caption and meta edit tagging information. }
    \label{fig:data_illustration}
\end{figure}

\begin{figure}[t]
    \centering
    \includegraphics[width=\linewidth]{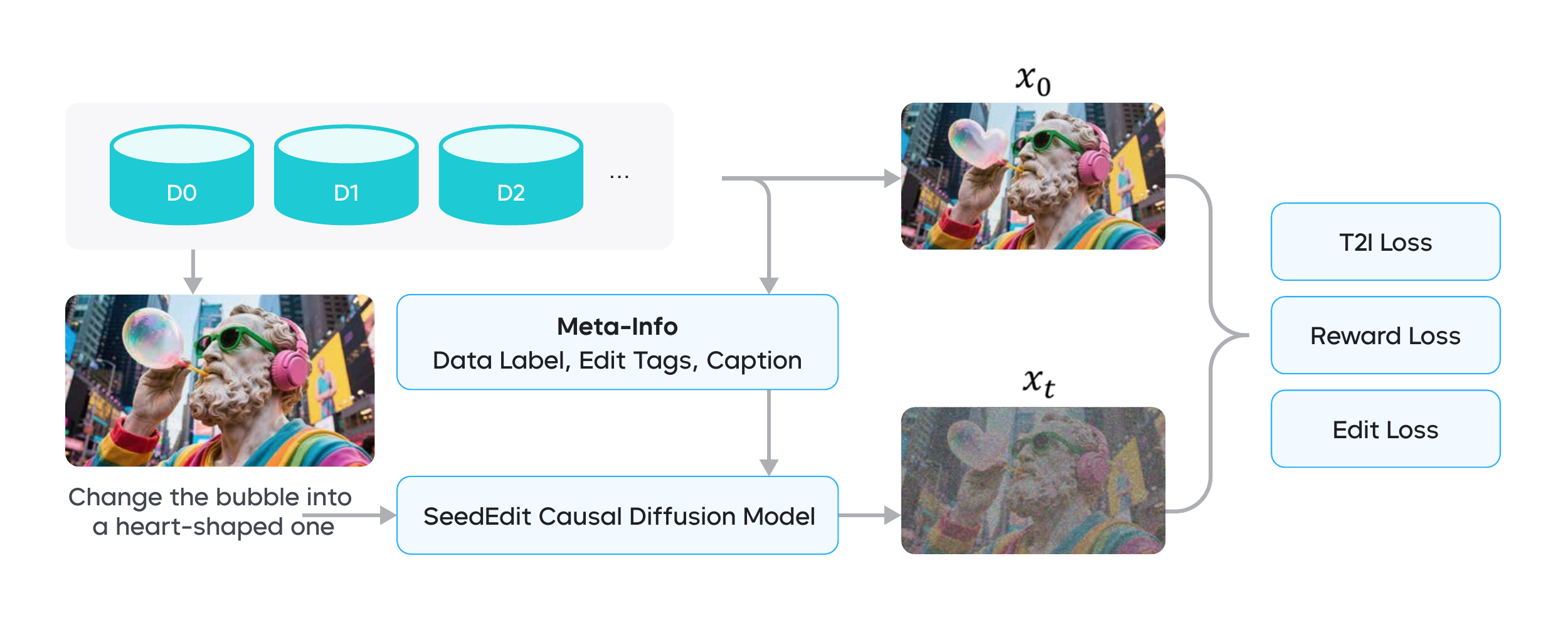}
    \caption{Training pipeline for SeedEdit3.0. We collect meta-info from multiple data sources and insert it in training by fusion multiple losses.}
    \label{fig:train_pipeline}
\end{figure}

In this section, we elaborate on the data curation strategies illustrated in Fig.~\ref{fig:data_illustration} and Fig.~\ref{fig:train_pipeline}, beginning with an introduction to several data sources which we paid attention to, followed by the data merging strategies with meta-information.

\subsubsection{Data Sources.} 
Specifically, we primarily collect data from the following sources, which help the diffusion model interleave the space of image editing for real and synthetic input output. 

\textbf{Synthesized dataset.}  From earlier work, e.g. HQEdit~\cite{hui2024hqedit}, we first noticed that modern diffusion models, such as DALLE3, exhibit strong in-context ability. Inspired by this orbservation, in our previous version, SeedEdit~\cite{shi2024seededit}, we extended this ability to our internal models by designing a novel pair data sampling strategy given the T2I and VLM models. In order to construct a general dataset with good coverage, our sampling includes both prompt sampling given the LLM/VLM~\cite{seed15vl} and noise sampling given the T2I~\cite{seedream3.0}. 

In this work, we further incorporate an importance sampling strategy that makes the sampled distribution aware of important and long-tail editing classes and subjects. This helps the synthesized data to achieve significantly broader coverage of different input and edit sample spaces. 

However, from observations in previous work~\cite{sheynin2024emu, shi2024seededit}, synthesized data have special biases towards the generated image domain, resulting in a performance gap between real images and synthetic data. In the following, we introduce how we handle and mitigate such a domain gap by carefully organizing the datasets.

\textbf{Editing specialists.} The first type of data that uses a real image as input comes from editor specialists, as also mentioned in recent work such as OmniEdit~\cite{wei2024omniedit}. In our internal community, there already exists a significant number of image editing specialists, such as those from ComfyUI~\cite{comfyui} workflows and pipelines, in-house specially optimized stylization, background modification, lighting adjustment, identity-aware DreamBooth, text editing, and more. These workflows typically take real images as input and produce outputs from well-designed generative models.

Therefore, we collaborated with our in-house image generation specialists to build multiple data-creation pipelines that well cover the design editing specialists. This synthesized dataset is particularly helpful for ensuring our dataset covers real-image input scenarios. 
Additionally, it also enables us to quickly address missing capabilities with our product design.
%Additionally, it also enables us to provide requests to the product team to quickly address missing capabilities.
% In addition, this can also help us to provide requests for product teams to quickly modify missing abilities. 

\textbf{Traditional Edit Operators.} To better support realistic and accurate image outputs after editing, we consider high-quality real image editing operations from traditional editing tools and software, such as lens blur, lighting adjustment, cropping, and template poster printing. As introduced in PromptFix~\cite{yu2024promptfixpromptfixphoto}, such types of datasets provide accurate loss directions in the real-image domain. Therefore, we also synthesized data from traditional editing operators,  where the edited images are based on multiple shots of a single item or through template-based editing operations, as shown in Fig.~\ref{fig:data_illustration}(b). In our experiments, we found that although these data cover a limited editing domain, they enable the model to produce realistic and accurate image rendering results.

\textbf{Video Frames and Multi Shorts} 
In addition to the aforementioned datasets, we recognize that large-scale and diverse real image data are crucial for improving the generalization ability of the editing model. Videos serve as a natural source of related pairs or groups of images, which can be captioned for image editing tasks. To sample such image-editing pairs from videos, we first randomly sample several key frames from each video clip. These key frames are then coarsely filtered based on CLIP image similarity and optical flow metrics. Finally, a VLM is applied to recapture and annotate the data, as described in Sect.~\ref{subsec:data_merge}.

% Data Prompt pipeline, we consider 
% \peng{peng's part} Same ID multiple scene from social media and products. 

\subsection{Data Merging}
\label{subsec:data_merge}
We propose a multi-granularity label strategy to effectively combine different sources of image editing data, i.e., data-level task label, text-level recaption, and pixel-level tagging, which we elaborate on in the following:

\paragraph{\textbf{Task Label.}}
We find that directly adding different sources of editing data to the original synthesized image pairs can lead to degraded performance. This degradation occurs because the data have very different editing styles. For example, the instruction ``change to Paris'' might imply a simple background replacement in traditional editing tasks, but it might also imply changing all pixels in the image in IP/ID preservation tasks, . Such diversity causes increased randomness in the test cases, most of which correspond to traditional editing scenarios. To address this, we distinguish between different data sources using task labels. High-quality data corresponding to traditional instruction-based editing are assigned a default editing label, which is also applied to all test inputs.

\paragraph{\textbf{Re-captioning.}}
We note that the main source of randomness introduced by diverse data sources is the ambiguity of the task condition, i.e., prompt description. Thus, another way to distinguish between different tasks while enabling knowledge transfer across them is to describe the tasks more clearly. In our data collection stage, many editing data contain incorrect or missing captions. For example, due to the biases in T2I models, synthesized prompt-to-prompt image pairs often include unintended changes that are not described in the original prompts. Similarly, video frames typically only have clip-level captions instead of inter-frame instructions. 

To address these issues, we design a novel recaptioning pipeline for image editing, where we decompose the task into two steps: (1) identifying all differences and similarities between the images, and (2) generating captions/instructions based on these differences. We find that this decomposed approach leads to improved accuracy with more details of the re-captioned descriptions.

\paragraph{\textbf{Tagging.}} To further improve the controllability of the editing models, we annotate the data with editing tags in addition to task labels and detailed prompts. These tags include local editing, face preservation, structure preservation, and style preservation, and are computed using VLMs or specialized models. During training, the editing model is conditioned on task labels, editing tags, and the re-captioned editing prompts.

To ensure balanced performance in bilingual settings, we perform prompt sampling and re-captioning using VLM~\cite{seed15vl} in both English and Chinese. In Fig.~\ref{fig:data_illustration}, we illustrate several collected examples for each category, along with their corresponding captions and tags. 

Finally, to fully leverage these datasets, we observe that all data can be trained with forward and backward editing operations after recaptioning, filtering, and alignment. This approach enables a good overall balance and well-covered data curation.

\subsection{Models}
\label{subsec:model}
In this section, we introduce our model, including model architecture and training strategies:

\subsubsection{Models Architecture}
Our model generally builds upon the architecture proposed in SeedEdit~\citep{shi2024seededit}, where a Vision-Language Model (VLM) at the bottom infers high-level semantic information from the image, and a causal diffusion network at the top reuses the diffusion process as an image encoder to capture fine-grained details. Between these components, a connector module is introduced to align the editing intent—such as task type and editing tag information—with the diffusion model, as discussed in Sect.~\ref{subsec:data_merge}.

In this work, we first replace the diffusion network from Seedream 2.0~\cite{seedream2.0} with Seedream 3.0~\cite{seedream3.0}, which can natively generate images at approximately 1024×1024 resolution without requiring any refiner. This upgrade significantly benefits the editing performance in terms of preserving input image details such as face and object identity. In addition, we leverage the model’s improved text-rendering capabilities for bilingual text and character-level editing.

\begin{figure}[t]
    \centering
    \centering
    \includegraphics[width=\linewidth]{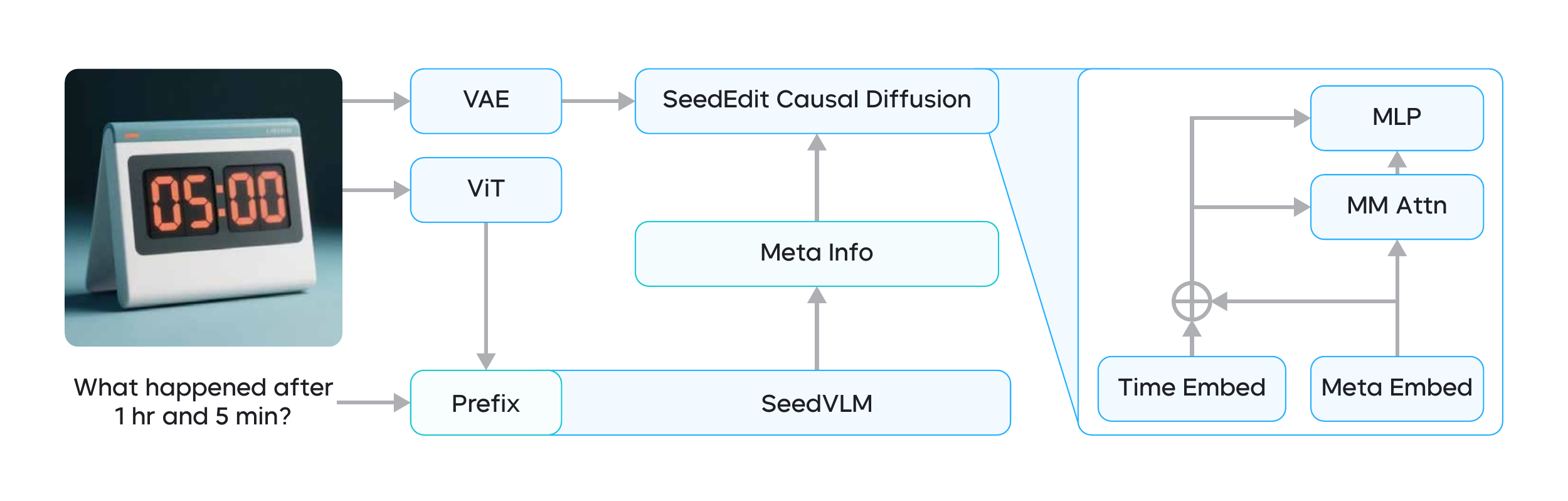}
    \caption{Model architecture with meta info embedding that connect VLM and the causal diffusion models.}
    \label{fig:arc_illustration}
\end{figure}

To more effectively combine task labels and tagging information from our labeled dataset, we introduce independent task embeddings for task label and tag injection. Compared with 
injection methods such as prompt-based ones~\cite{zhang2024hive}, we find that task embeddings enable the model to better distinguish between different dataset properties. Furthermore, classifier-free guidance (CFG) tricks can be optionally applied to further improve performance. Fig.~\ref{fig:arc_illustration} illustrates our model architecture, which can also be easily generalized to multimodal image generation tasks.

% \paragraph{\textbf{Edit Tags in Condition.}}
% \peng{peng finish}
% this is for training, show results with reestimated edit tags

\subsection{Model Training. }

To train this architecture, we adopt a multi-stage training strategy consisting of a pretraining stage that fuses all collected image pairs, and a fine-tuning stage that refines the outputs to stabilize editing performance.

\paragraph{\textbf{Multi-Aspect Ratio Training.}} 
Since our dataset in this version contains significantly more images with varying aspect ratios and resolutions, we modified the training pipeline to use NaViT~\cite{dehghani2023patch}, which supports batching images of different resolutions. In addition, we group images by resolution, enabling the model to progressively train from low to high resolution. For each resolution group, we dynamically adjust the maximum token length to maintain consistent batch sizes during training. This strategy helps preserve performance and retain information from previous stages.

During the fine-tuning stage, we re-sample a large amount of high-quality, high-resolution data from our curated datasets as the fine-tuning data. These samples are selected using a set of filter models and human filters together, to ensure both high quality and good coverage of editing classes.

\paragraph{\textbf{Diffusion with Reward Models.}}
In general, we adopt diffusion losses to finetune the model, which treats every part of the image equally important; however, certain attributes are especially high-value to users, such as face identity , some detailed structures and aesthetics, etc.

Therefore, we propose jointly training the model with a set of reward models that account for these attributes. Formally, let the rewards provided by these models be $R_i(\mathbf{x}_0, \mathbf{x}_1 | c)$, where $\mathbf{I}_0$ and $\mathbf{I}_1$ are input/output images, and $c$ is the condition indicating whether the reward should be considered. Our modified diffusion loss is defined as:

 \begin{equation}
 \label{eq:joint diffusion loss}
     L = \mathbb{E}_{t,q}
     \parallel
     \mathbf{v}_{\theta}\left(\mathbf{x}_1^{t}, t | c, \mathbf{x}_0 \right)
     - (\boldsymbol{\epsilon} - \mathbf{x}_1)\parallel_2^2 + \sum_i\lambda_iR_i(\mathbf{x}_0, \mathbf{x}_1^* | c, t)
 \end{equation} 
 Here, we also adopt the rectified flow matching~\cite{flux2024} as the diffusion loss. In addition, most of our rewards can only be calculated when the output image $\mathbf{x}_1^*$ can be reliably estimated at a given timestep $t$, and under the right instruction context $c$. For example, if the editing instruction requests a face change, there is no need to apply a reward for facial identity preservation.

One might consider using a unified model with paired image inputs; however, we find that current VLM-based models are not good at detail partition, resulting in lower performance compared to a set of expert reward models. We believe that as VLMs continue to improve in understanding image details, the reward models used in this work could eventually be merged and replaced.

\paragraph{\textbf{Joint training with T2I.}}
Last but not least, we notice that the quality of editing data is considerably lower than that of the best text-to-image (T2I) datasets, it is important to jointly train the model on both editing and T2I data, which brings two key benefits. First, by injecting high-quality, high-resolution images, we observe a significant improvement in the model’s editing ability on high-resolution images. Second, using T2I data helps preserve the model’s original T2I ability, which also contributes to better generalization in editing tasks.

\subsection{Inference Efficiency}
\subsubsection{Distillation} \label{sec: SD}
Our acceleration framework builds upon Hyper-SD~\cite{ren2025hyper} and RayFlow~\cite{shao2025rayflow}. We rethink the diffusion process by assigning each sample its own tailored generative path, rather than routing all examples through the same trajectory toward a fixed Gaussian prior. In traditional methods, all inputs are gradually transformed into isotropic Gaussian noise, leading to overlapping paths in probability space. These overlaps introduce additional randomness, weaken fine-grained control, and destabilize the reverse denoising process. In contrast, our approach assigns each sample a unique target distribution, greatly reducing path overlap and boosting both the stability of generation and the diversity of outputs.

\textbf{CFG Distillation.}
Classifier-Free Guidance (CFG) entails two network evaluations per time step-—one conditional and one unconditional--nearly doubling inference cost. To remedy this, we encode the guidance scale as a learnable embedding fused with the timestep encoding. Through targeted CFG distillation on this joint embedding, our model learns to deliver guided outputs in a single forward pass, achieving approximately two times faster inference while while preserving the ability to adjust guidance strength on demand.

\textbf{Unified Noise Reference.} 
To ensure smooth transitions throughout sampling, we employ a single noise reference vector predicted by a pre-trained network. This vector acts as a constant guide at each timestep, helping to align the denoising process over time. By maintaining a steady noise expectation, we reduce the total number of sampling steps without compromising fidelity. Our theoretical analysis further shows that this design maximizes the joint likelihood of the forward (data-to-noise) and reverse (noise-to-data) trajectories, resulting in stronger sampling performance and more faithful reconstructions.

\textbf{Adaptive Timestep Sampling.} 
We also streamline training by concentrating effort where it matters most. Conventional diffusion training samples timesteps uniformly at random, resulting in high variance in the loss and wasted computation on less informative intervals. To address this, we introduce an adaptive sampling strategy that concentrates on the most impactful timesteps. We combine the Stochastic Stein Discrepancy (SSD) criterion with a lightweight neural module that learns a data-driven timestep distribution. During training, this module identifies the timesteps that yield the greatest loss reduction, allowing more targeted updates. As a result, our method converges faster and utilizes computational resources more efficiently, significantly reducing the training cost.

\textbf{Few-Step, High-Fidelity Sampling.} 
Our framework supports very low-step sampling without compromising output quality. We adopt a tightly compressed denoising schedule that uses far fewer steps than standard baselines. Despite this compression, our method matches or outperforms approaches that require up to 75 function evaluations (NFE) across key metrics such as aesthetic quality, text–image alignment, and structural accuracy. These results demonstrate that our instance-aware trajectories and unified noise reference enable top-tier image synthesis with minimal computational overhead.

\subsubsection{Quantization and Overall Speedup}
Considering the architecture and scale of the DiT model, we optimize the performance of specific operators through techniques such as kernel fusion and memory access coalescing. As a result, the performance of certain operators more than doubles compared to their original implementations. Furthermore, we enhance performance and reduce memory usage through low-bit quantization of GEMM and Attention modules. On one hand, we propose an adaptive hybrid quantization approach to improve quantization accuracy. Specifically, we design an offline smoothing method to handle outliers in quantization layers. For sensitive layers in the model, we employ a search-based strategy to determine the optimal quantization granularity and scaling factors, which maximizes the quantization effectiveness. Finally, we fine-tune the model using post-training quantization to identify the optimal quantization parameters for each layer. On the other hand, we develop efficient quantized operators supporting various granularities and bit widths, which, when integrated with our quantization algorithms, achieve optimal performance. 
Excluding the VLM stage, our combined distillation and quantization pipeline delivers an 8× end-to-end inference speedup, reducing total runtime from approximately 64 s to 8 s.

%% file: sections/experiments.tex
\section{Experiments}
\label{sec:experiment}

\begin{figure}[t]
    \centering
    \centering
    \includegraphics[width=0.49\linewidth]{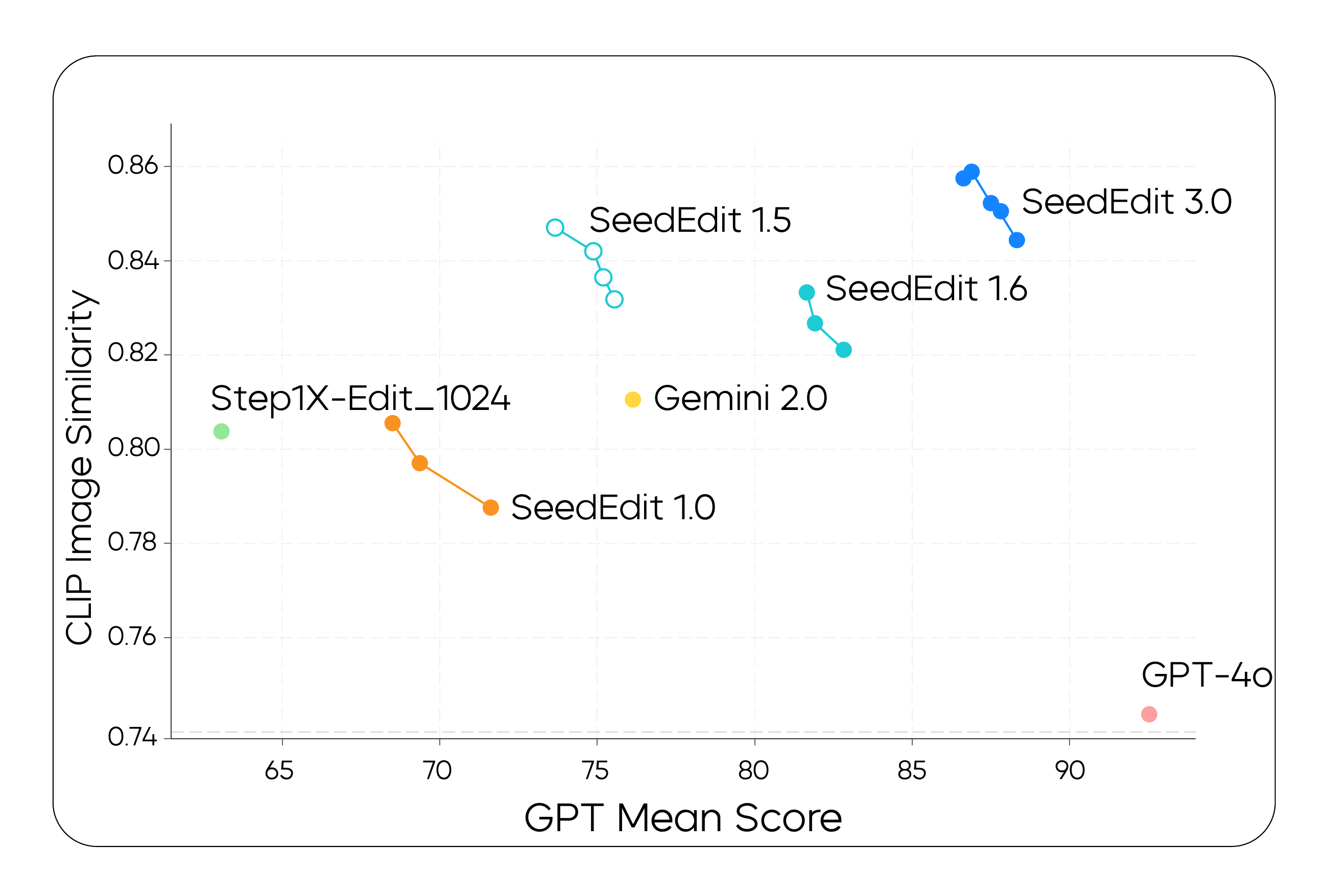}
    \includegraphics[width=0.49\linewidth]{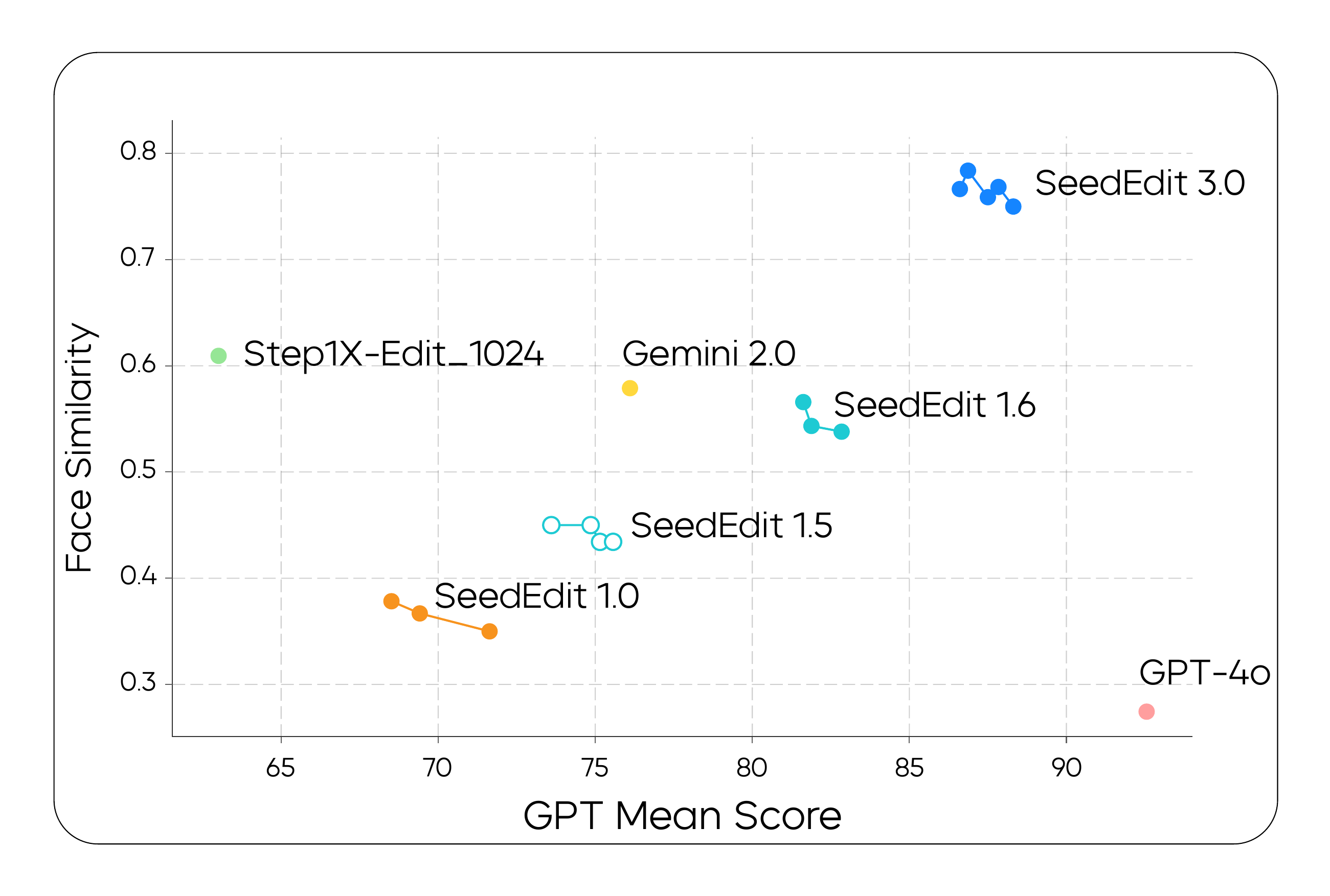}
    \caption{Quatitative comparisons with our previous versions and other SoTA methods. \textbf{Left:} GPT Mean Score vs. CLIP Image Similarity. \textbf{Right:} GPT Mean Score vs. Face Similarity. Different points for SeedEdit are obtained by changing the image CFG and text CFG, as proposed in~\cite{brooks2023instructpix2pix}, to observe its trade-off on image consistency and prompt following. } 
    \label{fig:quati}
\end{figure}

In this section, we elaborate on our setup of experiments, including evaluation sets and metrics.

\begin{figure}[!hbpt]
    \centering
    \includegraphics[width=\linewidth]{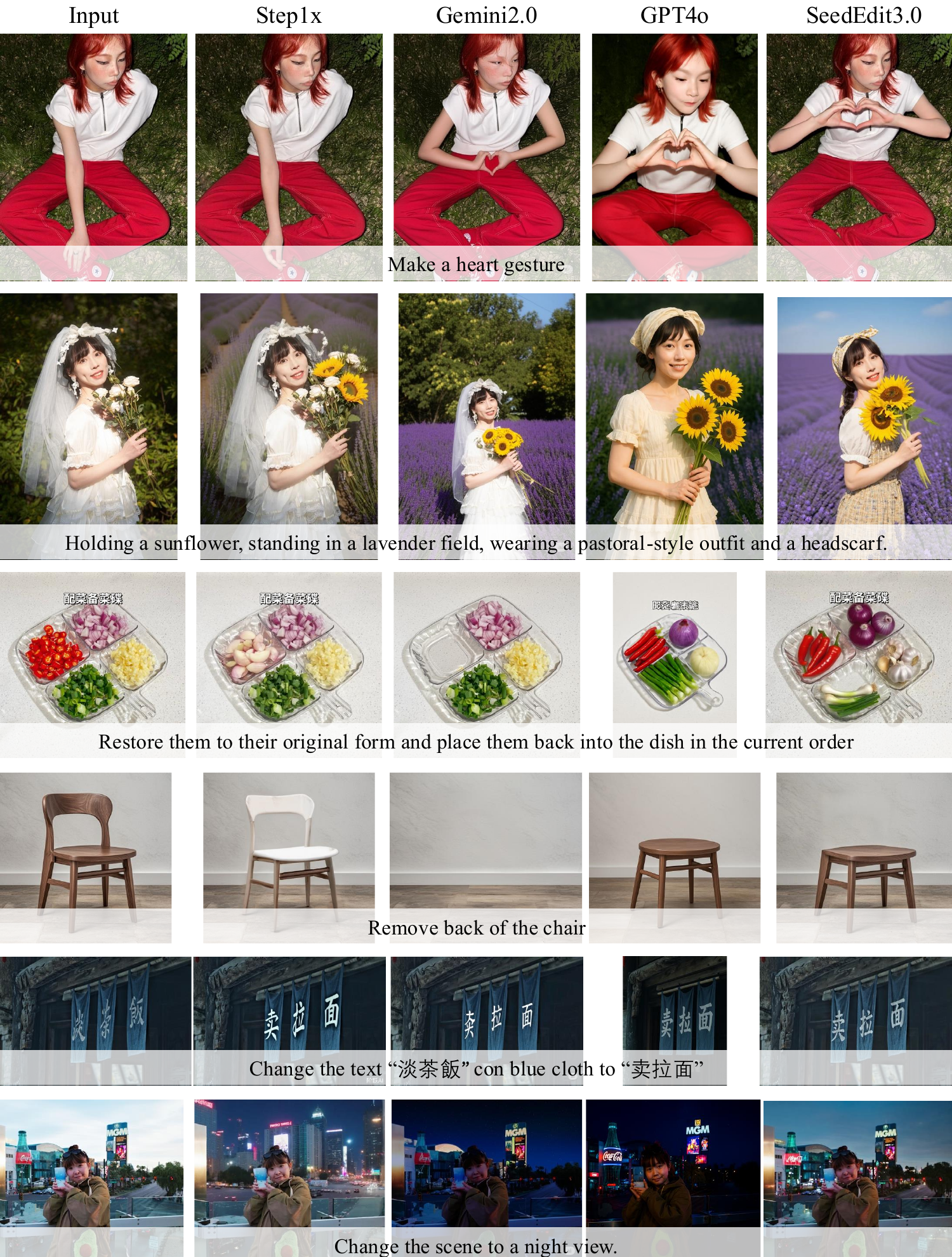}
    \caption{Qualitative comparisons. Notice the advantage of SeedEdit3.0 in face, object/human foreground and image detail preservation and alignment. }
    \label{fig:quali}
\end{figure}
%\subsection{Hello World}
\subsection{Evaluation.}
We first collected a few hundred testing images, based on both real and generated images. These test image sets include a wide range of editing operations. To be specific, in addition to common stylization, add, replace, and delete, we also include many instructive motions from camera, object shift, scene shot change, etc. which provide us a good guide on how well the model is performed in general user usage, rather than biased towards a few cases. 

For evaluation metrics, we consider the CLIP image simlairty and CLIP direction score metric proposed in InstructPix2Pix~\citep{brooks2023instructpix2pix}, and the GPT scores with the same GPT-4o model mentioned in HQEdit~\citep{hui2024hqedit} for quick machine-based evaluation. 
To make the evaluation more solid for product applications, we also adopt a set of human evaluations for the final quality check, shown in Fig.~\ref{fig:performance}, including a 0-5 scoring standard in three aspects: 1) instruction response, which evaluates whether the model responded to the instruction; 2) image consistency, whether the model preserved the identity after change; 3) image quality, which evaluates whether the model generates images with good quality without artifacts.  To summarize, we also provide satisfactory rates from 0-100, e.g. Usability Rate and Satisfaction Rate, as the percentage of satisfied edited images for a final summarization of model performance to benchmark different methods. Usability Rate means the results have minor non-satisfaction points (<3), and Satisfactory Rate means having 0 non-satisfaction points. In Fig.~\ref{fig:performance}, we set the max rate for Usability and Satisfaction to 60\% and 30\% respectively for better visualization due to our strict standards.  This metric aligns the standards from our T2I~\cite{seedream3.0} models and user feedback from our previous release versions, which has proven to be effective in online user performance evaluation.

\subsection{Comparisons}
 As shown in Fig.~\ref{fig:quati}, we show the quantitative comparison results of auto-machine evaluation with a few SoTA algorithms such as Step1x~\cite{liu2025step1x}, Gemini~\cite{gemini2.0} and GPT4o~\cite{hurst2024gpt}. The better models are located at the right top of the figure. 
To be fair on numbers, we run the open-source model with 1024 resolution for Step1x, and run others with their website chatting window with 4 times and choose the visually best one. For non-responded image queries, especially GPT4o and Gemini, we omit the score in evaluation. We have conducted extensive experiments with comparisons, and found that such evaluation metrics are well aligned with human feeling. 

We compare the current version with our multiple previous versions: SeedEdit1.0~\cite{shi2024seededit}, SeedEdit1.5 (by adding more data sources), and SeedEdit1.6 which were the data merging strategy and reward modeling added.
%
% Compared with our previous versions SeedEdit1.0~\cite{shi2024seededit}, in this report, we named adding more data sources as SeedEdit1.5, and adding the data merging strategy and reward modeling as SeedEdit1.6. 
 %
 This ablates different strategies as discussed in Sec.~\ref{subsec:data}.
 Our final model is represented by the yellow dots, namely SeedEdit3.0, which significantly improves over our previous versions and also outperforms other methods such as Gemini and Step1x in both metrics. For GPT-4o, it is located at the right bottom, which demonstrates that it has better prompt and instruction-following ability, while we find it has relatively weak image consistency, as demonstrated by CLIP image similarity and face similarity, which significantly impacts its human satisfaction rate as evaluated previously.

 The overall performance curve based on human evaluation has been illustrated in Fig.~\ref{fig:performance}, which shows that SeedEdit3.0 has the best trade-off across multiple metrics,  yielding the highest satisfaction rate for users. More importantly, we are comparably much faster, as it takes only 10-15s per image, compared to 50-60s per image for GPT-4o. 

Fig.~\ref{fig:quali} illustrates more comparison examples between our SeedEdit3.0 and SoTA models, where we further confirm the conclusion of our evaluation; we use results from single set of CFGs rather than picking from multiple ones. Additionally, we also notice a recent open source model: Step1X~\cite{liu2025step1x}, and find that it is difficult to compare directly with these commercial models, especially in real image quality, editing intention following, and understanding. 

\section{Conclusion}
In this report, we have introduced SeedEdit3.0, which significantly improves our previous versions in terms of real image performance, face/id preservation, text editing quality, prompt understanding, dynamic motion, etc. We presented an efficient data curation
pipeline that allows it to scale editing data  effectively; while a joint learning method was introduced to further enhance image consistency, which is particularly important for real-world applications. This results in a high-performance image editing system, which hopefully can be well used and adopted by all users to enrich their creativity.

%% file: sections/appendix.tex
\section{Other Contributors}
%\subsection{Hello World}
Here, we also thank many other team members who largely contributed to a successful deployment of the model, provided suggestions, and supported this work, including Yameng Li, Meng Guo for help with the data evaluation, Tianyu Zhao, Huafeng Kuang, Hao Li, Yawei Wen for model acceleration engineering, Haoshen Chen, Liang Li, Zuxi Liu, Bibo He for model deployment engineering.

\section{Ethical Claims}
The images presented in the paper are from our lisenced ones, and public license-free websites such as Unsplash and Pixabay. In addition, note that the technique proposed in this paper aims to facilitate the user's common tasks that are widely demanded in industry for ethical purposes. It SHOULD NOT be applied to unwanted scenarios such as generating violent and sexual content. It might also inherit the biases and limitations of T2I models. Therefore, we believe that the images or models synthesized using our approach should be carefully examined and presented as synthetic.

% Ordered first by area of contributions and then by the importance of each contributor. * indicates equal contributions. $\dagger$ indicates area lead in this paper. 

% \paragraph{\textbf{Research}} Peng Wang*^$\dagger$, Yichun Shi*, Xiaochen Lian, Zhonghua Zhai
% \paragraph{\textbf{Inference Efficiency}} Tianyu Zhao, Huafeng Kuang, Hao Li, Xin Xia, Xuefeng Xiao$\dagger$
% \paragraph{\textbf{Testing}} Yameng Li, Meng Guo, Liang Li
% \paragraph{\textbf{Engineer}} Haoshen Chen, Zuxi Liu, Bibo He^$\dagger$
% \paragraph{\textbf{Team Lead and Manager}} Weilin Huang, Jianchao Yang

% Core contributors indicate full-time efforts on this work, and contributors indicate members involved with partial time. * indicates equal contributions and $\dagger$ indicates the lead and manager of the team. 

% \paragraph{\textbf{Core Contributors}} Peng Wang*, Yichun Shi*, Xiaochen Lian, Zhonghua Zhai, Weilin Huang, Jianchao Yang

%% file: paper.bbl
\begin{thebibliography}{38}
\providecommand{\natexlab}[1]{#1}
\providecommand{\url}[1]{\texttt{#1}}
\expandafter\ifx\csname urlstyle\endcsname\relax
  \providecommand{\doi}[1]{doi: #1}\else
  \providecommand{\doi}{doi: \begingroup \urlstyle{rm}\Url}\fi

\bibitem[Brooks et~al.(2023)Brooks, Holynski, and Efros]{brooks2023instructpix2pix}
Tim Brooks, Aleksander Holynski, and Alexei~A Efros.
\newblock Instructpix2pix: Learning to follow image editing instructions.
\newblock In \emph{Proceedings of the IEEE/CVF conference on computer vision and pattern recognition}, pages 18392--18402, 2023.

\bibitem[Brown et~al.(2020)Brown, Mann, Ryder, Subbiah, Kaplan, Dhariwal, Neelakantan, Shyam, Sastry, Askell, et~al.]{brown2020language}
Tom Brown, Benjamin Mann, Nick Ryder, Melanie Subbiah, Jared~D Kaplan, Prafulla Dhariwal, Arvind Neelakantan, Pranav Shyam, Girish Sastry, Amanda Askell, et~al.
\newblock Language models are few-shot learners.
\newblock \emph{Advances in neural information processing systems}, 33:\penalty0 1877--1901, 2020.

\bibitem[Cao et~al.(2023)Cao, Wang, Qi, Shan, Qie, and Zheng]{cao2023masactrl}
Mingdeng Cao, Xintao Wang, Zhongang Qi, Ying Shan, Xiaohu Qie, and Yinqiang Zheng.
\newblock Masactrl: Tuning-free mutual self-attention control for consistent image synthesis and editing.
\newblock In \emph{Proceedings of the IEEE/CVF international conference on computer vision}, pages 22560--22570, 2023.

\bibitem[Chen et~al.(2025)Chen, Bai, Chai, Xie, Zhao, Vinci, Lin, and Chang]{chen2025multimodal}
Liang Chen, Shuai Bai, Wenhao Chai, Weichu Xie, Haozhe Zhao, Leon Vinci, Junyang Lin, and Baobao Chang.
\newblock Multimodal representation alignment for image generation: Text-image interleaved control is easier than you think.
\newblock \emph{arXiv preprint arXiv:2502.20172}, 2025.

\bibitem[comfyanonymous et~al.(2023)]{comfyui}
comfyanonymous et~al.
\newblock Comfyui: The most powerful and modular stable diffusion gui.
\newblock \url{https://github.com/comfyanonymous/ComfyUI}, 2023.
\newblock Accessed: May 15, 2025.

\bibitem[Dehghani et~al.(2023)Dehghani, Mustafa, Djolonga, Heek, Minderer, Caron, Steiner, Puigcerver, Geirhos, Alabdulmohsin, et~al.]{dehghani2023patch}
Mostafa Dehghani, Basil Mustafa, Josip Djolonga, Jonathan Heek, Matthias Minderer, Mathilde Caron, Andreas Steiner, Joan Puigcerver, Robert Geirhos, Ibrahim~M Alabdulmohsin, et~al.
\newblock Patch n’pack: Navit, a vision transformer for any aspect ratio and resolution.
\newblock \emph{Advances in Neural Information Processing Systems}, 36:\penalty0 2252--2274, 2023.

\bibitem[Feng et~al.(2025)Feng, Ma, Wang, Qi, Chen, Chen, and Wang]{feng2025dit4edit}
Kunyu Feng, Yue Ma, Bingyuan Wang, Chenyang Qi, Haozhe Chen, Qifeng Chen, and Zeyu Wang.
\newblock Dit4edit: Diffusion transformer for image editing.
\newblock In \emph{Proceedings of the AAAI Conference on Artificial Intelligence}, volume~39, pages 2969--2977, 2025.

\bibitem[Gemini2(2024)]{gemini2.0}
Google Gemini2.
\newblock Experiment with gemini 2.0 flash native image generation, 2024.
\newblock \url{https://developers.googleblog.com/en/experiment-with-gemini-20-flash-native-image-generation/}.

\bibitem[Hertz et~al.(2022)Hertz, Mokady, Tenenbaum, Aberman, Pritch, and Cohen-Or]{hertz2022prompt}
Amir Hertz, Ron Mokady, Jay Tenenbaum, Kfir Aberman, Yael Pritch, and Daniel Cohen-Or.
\newblock Prompt-to-prompt image editing with cross attention control.
\newblock \emph{arXiv preprint arXiv:2208.01626}, 2022.

\bibitem[Huang et~al.(2024)Huang, Xie, Wang, Yuan, Cun, Ge, Zhou, Dong, Huang, Zhang, et~al.]{huang2024smartedit}
Yuzhou Huang, Liangbin Xie, Xintao Wang, Ziyang Yuan, Xiaodong Cun, Yixiao Ge, Jiantao Zhou, Chao Dong, Rui Huang, Ruimao Zhang, et~al.
\newblock Smartedit: Exploring complex instruction-based image editing with multimodal large language models.
\newblock In \emph{Proceedings of the IEEE/CVF Conference on Computer Vision and Pattern Recognition}, pages 8362--8371, 2024.

\bibitem[Hui et~al.(2024)Hui, Yang, Zhao, Shi, Weng, Wang, Zhou, and Xie]{hui2024hqedit}
Mude Hui, Siwei Yang, Bingchen Zhao, Yichun Shi, Heng Weng, Peng Wang, Yuyin Zhou, and Cihang Xie.
\newblock Hq-edit: Ahigh-quality dataset for instruc-tion based image editing.
\newblock \emph{arXiv preprint arXiv:2404.09990}, 2024.

\bibitem[Hurst et~al.(2024)Hurst, Lerer, Goucher, Perelman, Ramesh, Clark, Ostrow, Welihinda, Hayes, Radford, et~al.]{hurst2024gpt}
Aaron Hurst, Adam Lerer, Adam~P Goucher, Adam Perelman, Aditya Ramesh, Aidan Clark, AJ~Ostrow, Akila Welihinda, Alan Hayes, Alec Radford, et~al.
\newblock Gpt-4o system card.
\newblock \emph{arXiv preprint arXiv:2410.21276}, 2024.

\bibitem[Labs(2024)]{flux2024}
Black~Forest Labs.
\newblock Flux.
\newblock \url{https://github.com/black-forest-labs/flux}, 2024.

\bibitem[Liao et~al.(2025)Liao, Liu, Wang, Luo, Zhang, Zhao, Wu, Li, Tian, and Huang]{liao2025mogao}
Chao Liao, Liyang Liu, Xun Wang, Zhengxiong Luo, Xinyu Zhang, Wenliang Zhao, Jie Wu, Liang Li, Zhi Tian, and Weilin Huang.
\newblock Mogao: An omni foundation model for interleaved multi-modal generation.
\newblock \emph{arXiv preprint arXiv:2505.05472}, 2025.

\bibitem[Lin et~al.(2024)Lin, Chen, Wang, An, Wang, Tian, Liu, Dai, Wang, and Wang]{lin2024schedule}
Haonan Lin, Yan Chen, Jiahao Wang, Wenbin An, Mengmeng Wang, Feng Tian, Yong Liu, Guang Dai, Jingdong Wang, and Qianying Wang.
\newblock Schedule your edit: A simple yet effective diffusion noise schedule for image editing.
\newblock \emph{Advances in Neural Information Processing Systems}, 37:\penalty0 115712--115756, 2024.

\bibitem[Liu et~al.(2025)Liu, Han, Xing, Yin, Wang, Cheng, Liao, Wang, Fu, Han, et~al.]{liu2025step1x}
Shiyu Liu, Yucheng Han, Peng Xing, Fukun Yin, Rui Wang, Wei Cheng, Jiaqi Liao, Yingming Wang, Honghao Fu, Chunrui Han, et~al.
\newblock Step1x-edit: A practical framework for general image editing.
\newblock \emph{arXiv preprint arXiv:2504.17761}, 2025.

\bibitem[Mokady et~al.(2023)Mokady, Hertz, Aberman, Pritch, and Cohen-Or]{mokady2023null}
Ron Mokady, Amir Hertz, Kfir Aberman, Yael Pritch, and Daniel Cohen-Or.
\newblock Null-text inversion for editing real images using guided diffusion models.
\newblock In \emph{Proceedings of the IEEE/CVF conference on computer vision and pattern recognition}, pages 6038--6047, 2023.

\bibitem[Nie et~al.(2023)Nie, Guo, Lu, Zhou, Zheng, and Li]{nie2023blessing}
Shen Nie, Hanzhong~Allan Guo, Cheng Lu, Yuhao Zhou, Chenyu Zheng, and Chongxuan Li.
\newblock The blessing of randomness: Sde beats ode in general diffusion-based image editing.
\newblock \emph{arXiv preprint arXiv:2311.01410}, 2023.

\bibitem[OpenAI(2023)]{dalle3}
OpenAI.
\newblock Dall·e 3 system card, 2023.
\newblock \url{https://cdn.openai.com/papers/DALL_E_3_System_Card.pdf}.

\bibitem[Pan et~al.(2025)Pan, Shukla, Singh, Zhao, Mishra, Wang, Xu, Chen, Li, Juefei-Xu, et~al.]{pan2025transfer}
Xichen Pan, Satya~Narayan Shukla, Aashu Singh, Zhuokai Zhao, Shlok~Kumar Mishra, Jialiang Wang, Zhiyang Xu, Jiuhai Chen, Kunpeng Li, Felix Juefei-Xu, et~al.
\newblock Transfer between modalities with metaqueries.
\newblock \emph{arXiv preprint arXiv:2504.06256}, 2025.

\bibitem[Ren et~al.(2025)Ren, Xia, Lu, Zhang, Wu, Xie, Wang, and Xiao]{ren2025hyper}
Yuxi Ren, Xin Xia, Yanzuo Lu, Jiacheng Zhang, Jie Wu, Pan Xie, Xing Wang, and Xuefeng Xiao.
\newblock Hyper-sd: Trajectory segmented consistency model for efficient image synthesis.
\newblock \emph{Advances in Neural Information Processing Systems}, 37:\penalty0 117340--117362, 2025.

\bibitem[Seed Vision T2I~Team(2025)]{seedream3.0}
ByteDance Seed Vision T2I~Team.
\newblock Seedream 3.0 technical report.
\newblock \emph{arXiv preprint arXiv:2504.11346}, 2025.

\bibitem[Seed Vision~Team(2025)]{seedream2.0}
ByteDance Seed Vision~Team.
\newblock Seedream 2.0: A native chinese-english bilingual image generation foundation model.
\newblock \emph{arXiv preprint arXiv:2503.07703}, 2025.

\bibitem[Seed Vision Understanding~Team(2025)]{seed15vl}
Bytedance Seed Vision Understanding~Team.
\newblock Seed1.5-vl technical report, 2025.
\newblock URL \url{https://arxiv.org/abs/2505.07062}.

\bibitem[Shao et~al.(2025)Shao, Xia, Yang, Ren, Wang, and Xiao]{shao2025rayflow}
Huiyang Shao, Xin Xia, Yuhong Yang, Yuxi Ren, Xing Wang, and Xuefeng Xiao.
\newblock Rayflow: Instance-aware diffusion acceleration via adaptive flow trajectories.
\newblock \emph{arXiv preprint arXiv:2503.07699}, 2025.

\bibitem[Sheynin et~al.(2024)Sheynin, Polyak, Singer, Kirstain, Zohar, Ashual, Parikh, and Taigman]{sheynin2024emu}
Shelly Sheynin, Adam Polyak, Uriel Singer, Yuval Kirstain, Amit Zohar, Oron Ashual, Devi Parikh, and Yaniv Taigman.
\newblock Emu edit: Precise image editing via recognition and generation tasks.
\newblock In \emph{Proceedings of the IEEE/CVF Conference on Computer Vision and Pattern Recognition}, pages 8871--8879, 2024.

\bibitem[Shi et~al.(2024)Shi, Wang, and Huang]{shi2024seededit}
Yichun Shi, Peng Wang, and Weilin Huang.
\newblock Seededit: Align image re-generation to image editing.
\newblock \emph{arXiv preprint arXiv:2411.06686}, 2024.

\bibitem[Tumanyan et~al.(2023)Tumanyan, Geyer, Bagon, and Dekel]{tumanyan2023plug}
Narek Tumanyan, Michal Geyer, Shai Bagon, and Tali Dekel.
\newblock Plug-and-play diffusion features for text-driven image-to-image translation.
\newblock In \emph{Proceedings of the IEEE/CVF Conference on Computer Vision and Pattern Recognition}, pages 1921--1930, 2023.

\bibitem[Wang et~al.(2023)Wang, Zhang, Birsak, and Wonka]{wang2023instructedit}
Qian Wang, Biao Zhang, Michael Birsak, and Peter Wonka.
\newblock Instructedit: Improving automatic masks for diffusion-based image editing with user instructions.
\newblock \emph{arXiv preprint arXiv:2305.18047}, 2023.

\bibitem[Wei et~al.(2024)Wei, Xiong, Ren, Du, Zhang, and Chen]{wei2024omniedit}
Cong Wei, Zheyang Xiong, Weiming Ren, Xeron Du, Ge~Zhang, and Wenhu Chen.
\newblock Omniedit: Building image editing generalist models through specialist supervision.
\newblock In \emph{The Thirteenth International Conference on Learning Representations}, 2024.

\bibitem[Xiao et~al.(2024)Xiao, Wang, Zhou, Yuan, Xing, Yan, Li, Wang, Huang, and Liu]{xiao2024omnigen}
Shitao Xiao, Yueze Wang, Junjie Zhou, Huaying Yuan, Xingrun Xing, Ruiran Yan, Chaofan Li, Shuting Wang, Tiejun Huang, and Zheng Liu.
\newblock Omnigen: Unified image generation.
\newblock \emph{arXiv preprint arXiv:2409.11340}, 2024.

\bibitem[Yang et~al.(2024)Yang, Yu, Meng, Xu, Ermon, and Cui]{yang2024mastering}
Ling Yang, Zhaochen Yu, Chenlin Meng, Minkai Xu, Stefano Ermon, and Bin Cui.
\newblock Mastering text-to-image diffusion: Recaptioning, planning, and generating with multimodal llms.
\newblock In \emph{Forty-first International Conference on Machine Learning}, 2024.

\bibitem[Yu et~al.(2024{\natexlab{a}})Yu, Chow, Yue, Pan, Wu, Wan, Li, Tang, Zhang, and Zhuang]{yu2024anyedit}
Qifan Yu, Wei Chow, Zhongqi Yue, Kaihang Pan, Yang Wu, Xiaoyang Wan, Juncheng Li, Siliang Tang, Hanwang Zhang, and Yueting Zhuang.
\newblock Anyedit: Mastering unified high-quality image editing for any idea.
\newblock \emph{arXiv preprint arXiv:2411.15738}, 2024{\natexlab{a}}.

\bibitem[Yu et~al.(2024{\natexlab{b}})Yu, Zeng, Hua, Fu, and Luo]{yu2024promptfixpromptfixphoto}
Yongsheng Yu, Ziyun Zeng, Hang Hua, Jianlong Fu, and Jiebo Luo.
\newblock Promptfix: You prompt and we fix the photo, 2024{\natexlab{b}}.
\newblock URL \url{https://arxiv.org/abs/2405.16785}.

\bibitem[Zhang et~al.(2023)Zhang, Mo, Chen, Sun, and Su]{zhang2023magicbrush}
Kai Zhang, Lingbo Mo, Wenhu Chen, Huan Sun, and Yu~Su.
\newblock Magicbrush: A manually annotated dataset for instruction-guided image editing.
\newblock \emph{Advances in Neural Information Processing Systems}, 36:\penalty0 31428--31449, 2023.

\bibitem[Zhang et~al.(2024)Zhang, Yang, Feng, Qin, Chen, Yu, Chen, Wang, Savarese, Ermon, et~al.]{zhang2024hive}
Shu Zhang, Xinyi Yang, Yihao Feng, Can Qin, Chia-Chih Chen, Ning Yu, Zeyuan Chen, Huan Wang, Silvio Savarese, Stefano Ermon, et~al.
\newblock Hive: Harnessing human feedback for instructional visual editing.
\newblock In \emph{Proceedings of the IEEE/CVF Conference on Computer Vision and Pattern Recognition}, pages 9026--9036, 2024.

\bibitem[Zhao et~al.(2024)Zhao, Ma, Chen, Si, Wu, An, Yu, Zhang, Li, and Chang]{zhao2024ultraedit}
Haozhe Zhao, Xiaojian~Shawn Ma, Liang Chen, Shuzheng Si, Rujie Wu, Kaikai An, Peiyu Yu, Minjia Zhang, Qing Li, and Baobao Chang.
\newblock Ultraedit: Instruction-based fine-grained image editing at scale.
\newblock \emph{Advances in Neural Information Processing Systems}, 37:\penalty0 3058--3093, 2024.

\bibitem[Zhou et~al.(2024)Zhou, Yu, Babu, Tirumala, Yasunaga, Shamis, Kahn, Ma, Zettlemoyer, and Levy]{zhou2024transfusion}
Chunting Zhou, Lili Yu, Arun Babu, Kushal Tirumala, Michihiro Yasunaga, Leonid Shamis, Jacob Kahn, Xuezhe Ma, Luke Zettlemoyer, and Omer Levy.
\newblock Transfusion: Predict the next token and diffuse images with one multi-modal model.
\newblock \emph{arXiv preprint arXiv:2408.11039}, 2024.

\end{thebibliography}
